\documentclass[final]{article}

\usepackage{neurips_2024}

\pdfoutput=1

\usepackage[utf8]{inputenc} 
\usepackage[T1]{fontenc}    
\usepackage{hyperref}       
\usepackage{url}            
\usepackage{booktabs}       
\usepackage{amsfonts}       
\usepackage{nicefrac}       
\usepackage{microtype}      
\usepackage{xcolor}         
\usepackage{enumitem}

\usepackage{lineno}
\nolinenumbers 

\usepackage{multirow}

\usepackage{graphicx}
\usepackage{float}

\title{SAM-MPA: Applying SAM to Few-shot Medical Image Segmentation using Mask Propagation and Auto-prompting}

\makeatletter  
\renewcommand\@makefntext[1]{%
    \parindent 1em%
    \noindent  
    \hb@xt@1.8em{\hss\@makefnmark}#1}  
\makeatother  

\author{  
    Jie Xu \quad Xiaokang Li \quad Chengyu Yue \quad Chen Ma \quad Yuanyuan Wang\textsuperscript{*} \quad Yi Guo\textsuperscript{*} \\
    School of Information Science and Engineering, Fudan University, Shanghai, China \\
    \texttt{xujie23@m.fudan.edu.cn,\,lixiaokang@fudan.edu.cn,\,cyyue22@m.fudan.edu.cn} \\
    \texttt{cma24@m.fudan.edu.cn,\,yywang@fudan.edu.cn,\,guoyi@fudan.edu.cn}  
}  

\begin{document}

\footnotetext{\textsuperscript{*}Corresponding author}

\maketitle

\begin{abstract}
Medical image segmentation often faces the challenge of prohibitively expensive annotation costs. While few-shot learning offers a promising solution to allevi- ate this burden, conventional approaches still rely heavily on pre-training with large volumes of labeled data from known categories. To address this issue, we propose leveraging the Segment Anything Model (SAM), pre-trained on over 1 billion masks, thus circumventing the need for extensive domain-specific annotated data. In light of this, we developed SAM-MPA, an innovative \textbf{SAM}-based framework for few-shot medical image segmentation using \textbf{M}ask \textbf{P}ropagation-based \textbf{A}uto-prompting. Initially, we employ k-centroid clustering to select the most representative examples for labelling to construct the support set. These annotated examples are registered to other images yielding deformation fields that facilitate the propagation of the mask knowledge to obtain coarse masks across the dataset.  Subsequently, we automatically generate visual prompts based on the region and boundary expansion of the coarse mask, including points, box and a coarse mask. Finally, we can obtain the segmentation predictions by inputting these prompts into SAM and refine the results by post refinement module. We validate the performance of the proposed framework through extensive experiments conducted on two medical image datasets with different modalities. Our method achieves Dices of 74.53\%, 94.36\% on Breast US, Chest X-ray, respectively. Experimental results substantiate that SAM-MPA yields high-accuracy segmentations within 10 labeled examples, outperforming other state-of-the-art few-shot auto-segmentation methods. Our method enables the customization of SAM for any medical image dataset with a small number of labeled examples.
\end{abstract}

\section{Introduction}

Accurate image segmentation plays a pivotal role in medical image analysis and assisted diagnosis \cite{wang2022medical}. It involves the segmentation of tissues, organs, and lesions in medical images across various imaging modalities. Typically, these segmentation labels are painstakingly acquired through manual outlining by experienced doctors, utilizing specialized annotation tools like ITK-SNAP\cite{yushkevich2016itk}. Getting a large amount of labeled data is difficult due to the time-consuming and expensive process, as well as medical ethics concerns. Consequently, many deep learning models encounter obstacles in effectively training when confronted with limited data scenarios.

Addressing the challenge of limited labeled data availability, researchers have extensively explored few-shot segmentation (FSS) methods in the context of medical imaging\cite{song2023comprehensive}. FSS aims to delineate the target in a query image with the assistance of a small number of pixel-wise annotated support images. In general, FSS techniques rely on meta-learning approaches built upon prototype networks\cite{snell2017prototypical}.
These methods typically involve a two-step process: creating representative prototypes from known classes, then using these to segment new, unseen images. It usually requires substantial labeled training data from known object categories in the first step. However, in medical imaging, such extensive annotations are costly and time-consuming. Therefore, we need to develop new few-shot segmentation methods that do not require large amounts of labeled medical data.

In response to the issue, the Segment Anything Model (SAM\cite{kirillov2023segment}) has recently emerged as a promising approach. Trained on over 1 billion masks, SAM demonstrates a powerful generalized few shot segmentation capability, able to segment any subject using visual prompts, even those hasn't seen before. Several studies have already applied SAM to medical imaging, yielding promising segmentation results\cite{cheng2023sam,lin2023samus,ma2024segment,wu2023medical,zhang2023customized}. They primarily focus on fine-tuning SAM and manually providing prompts. However, there are two significant challenges. (1) Fine-tuning methods typically require anywhere from hundreds\cite{lin2023samus,wu2023medical,zhang2023customized} to millions\cite{cheng2023sam,ma2024segment} of masked training images. In contrast, FSS aims to achieve satisfactory segmentation performance on medical images using only a handful of masked samples (typically 1, 5, or 10 instances). (2) The practice of manually supplying prompts is labor-intensive and highly inefficient. Given these considerations, we propose to explore the application of SAM for FSS using automated prompts in medical image analysis.

\begin{figure}[H]
\centering
\includegraphics[width=\textwidth]{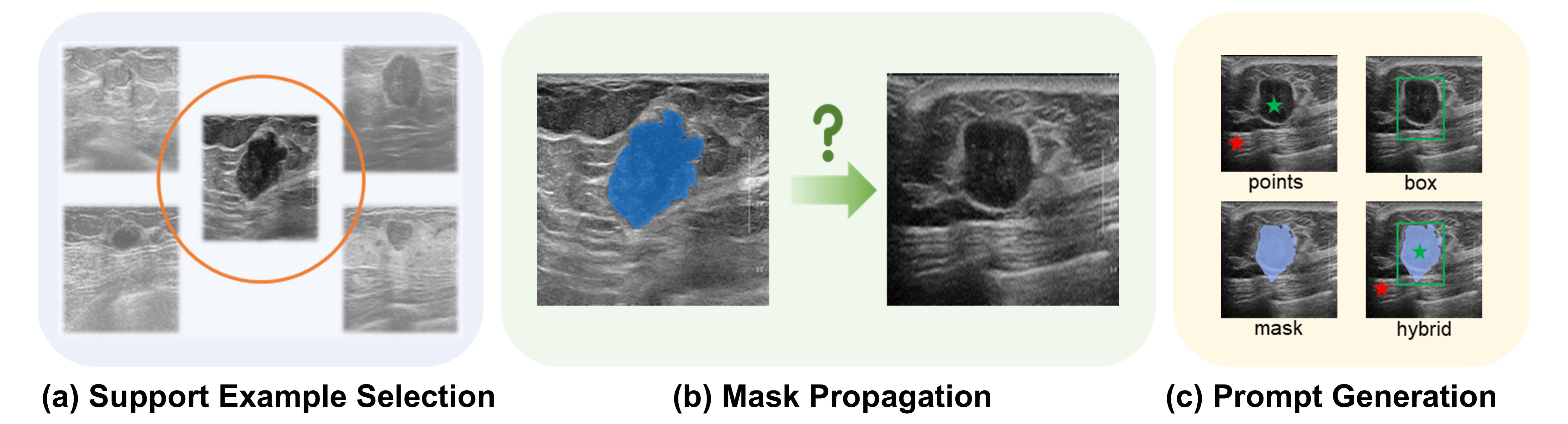}
\caption{Three key challenges in adapting SAM to few shot medical segmentation}
\label{fig1}
\end{figure}

Based on existing research, we have identified three key challenges when applying SAM to few-shot medical segmentation, as illustrated in Fig. \ref{fig1}: (1) Selection of appropriate labeled images as support examples. The choice will significantly impact the performance of the foundation visual model on downstream tasks\cite{zhang2023makes}. (2) Effective propagation of mask knowledge from support images to query images. This is the core challenge in few-shot learning scenarios\cite{wang2020generalizing}, as it demands the model to extract mask-relevant features from a minimal support set and accurately transefer prior knowledge to the diverse query set. (3) Generation of high-quality prompts. Well-crafted prompts are crucial for maximizing SAM's potential\cite{huang2024segment}.

To tackle these problems, we introduce SAM-MPA, a novel framework that enhances the \textbf{SAM} for few-shot medical image segmentation by integrating \textbf{M}ask \textbf{P}ropogation and \textbf{A}utomatic prompt generation techniques. First, we cluster samples based on their intrinsic structure and select the most representative instance from each cluster to form the support image set. Second, we perform registration between support and query images, utilizing the resulting deformation field to transfer mask knowledge. Third, leveraging the deformed masks, we automatically generate prompts including foreground points, background points, and bounding boxes.

Our contributions can be summarized as follows:
\begin{enumerate}[leftmargin=*, align=left]  
  \item We propose SAM-MPA, an innovative framework that successfully adapts SAM for few-shot medical image segmentation through automatic prompting. It integrates support sample selection, mask propagation, and automatic prompt generation, achieving high-performance segmentation results with only 1-10 labeled samples.  
  
  \item To address existing challenges, we introduce three innovative strategies: a clustering-based support sample selection, an unsupervised registration-driven mask propagation, and an automatic high-quality prompt generation. They respectively provide a representative support set, facilitate efficient knowledge transfer between support and query images, and generate optimal prompts for SAM.  
  
  \item Our method achieves Dices of 82.07\%, 94.13\% on Breast US and Chest X-ray datasets, respectively. Experimental results demonstrate the effectiveness of our method, showcasing superior segmentation accuracy while requiring minimal labeled data across different medical imaging modality.
\end{enumerate}

\section{Related Work}

\subsection{Few-shot Medical Image Segmentation}
Few-shot segmentation in natural image typically rely on extensive pre-training using labeled data from known semantic categories. This approach has influenced medical imaging, leading to the development of methods such as SE-Net\cite{roy2020squeeze}, MRrNet\cite{feng2021interactive}, and GCN-DE\cite{sun2022few}, which utilize known category data to train efficient dual-branch interaction models. However, obtaining large-scale pre-labeled datasets in the medical domain is often impractical. To address this challenge, some researchers have turned to self-supervised learning techniques\cite{cheng2024few,hansen2022anomaly,ouyang2020self,wang2022few}. These approaches generate pseudo-labels through superpixel\cite{felzenszwalb2004efficient}, enabling self-supervised pre-training. While this method reduces annotation costs, it still faces several limitations: (1) the necessity of a pre-training phase. (2) the potential inaccuracies in superpixel generation. (3) the requirement for large-scale datasets to achieve effective self-supervised pre-training, which is often unfeasible in medical imaging. In contrast, with its publicly available pre-trained weights and extensive training on a vast dataset, SAM eliminates the need for additional domain-specific pre-training or large-scale labeled datasets. Therefore, we can consider SAM as an effective soultion for few-shot medical image segmentation.




\subsection{A revisit of SAM}
\label{sec2.2}
SAM is a universal image segmentation model trained on over 11 million images and 1 billion masks. It consists of three main components: (1) an image encoder, built on a refined Vision Transformer\cite{dosovitskiy2020image}, which extracts robust image features. (2) a prompt encoder that converts diverse user prompts—ranging from sparse (points, boxes) to dense (masks)—into a unified feature space. (3) a light mask decoder that synthesizes inputs from two encoders to generate precise segmentations. SAM’s primary distinction from traditional models lies in its ability to flexibly accept various forms of prompts. These prompts offer indicative information about the target area, enabling SAM to produce accurate results for new objects, thus exhibiting powerful generalization capabilities. Consequently, when adapting SAM to new domains, providing high-quality prompts becomes one of the most crucial tasks.

\subsection{Adapting SAM to Medical Image Segmentation}
The remarkable performance of SAM on natural images has led researchers to explore its capabilities in medical image segmentation. Adaptation methods for SAM in this domain can be categorized into fine-tuning and auto-prompting strategies\cite{zhang2024segment}. Fine-tuning approaches include full fine-tuning (e.g., MedSAM\cite{ma2024segment}) and parameter-efficient techniques (such as Med-SA\cite{wu2023medical}, SAM-Med2D\cite{cheng2023sam}, SAMed\cite{zhang2023customized}, and AdaptiveSAM\cite{paranjape2024adaptivesam}). Full fine-tuning updates all model parameters using large annotated datasets, while parameter-efficient methods adjust only some parameters to conserve resources. These fine-tuning approaches have drawbacks: they require substantial computational resources, incur labeling costs(MedSAM used over 1.5 million image-mask pairs), and rely on manually provided prompts in training data. Even parameter-efficient methods need hundreds to thousands of annotated images. Auto-prompting adaptation strategies aim to reduce reliance on manual prompts. For example, Pandey et al. used YOLOv8 to generate boundary boxes as prompts\cite{pandey2023comprehensive}, but this still requires 100 masked images for training. PerSAM\cite{zhang2023personalize} achieves one-shot segmentation for specific categories in natural images using a single annotated image, employing feature matching to generate point prompts. Such approach may be less effective in medical imaging where target structures and surrounding tissues often have minimal visual differences.

\section{Method}
\subsection{Problem Definition}
Let $D=\{d_1,d_2,\cdots,d_N\}$ is an unlabeled dataset which contains $N$ samples. In the few-shot scenario, we define the support set as $D_S=\{(s_1, m_1), (s_2, m_2),\cdots, (c_K, m_K)\}$, where $m_i$ represents the mask corresponding to each image. The query set is denoted as $D_Q=\{q_1,q_2,\cdots,q_{N-K}\}$. Typical few-shot settings use $K$ values of 1, 5, or 10. Our goal is to leverage the mask information in $D_S$ to achieve high-precision segmentation with SAM on $D_Q$.

\begin{figure}[H]
\centering
\includegraphics[width=\textwidth]{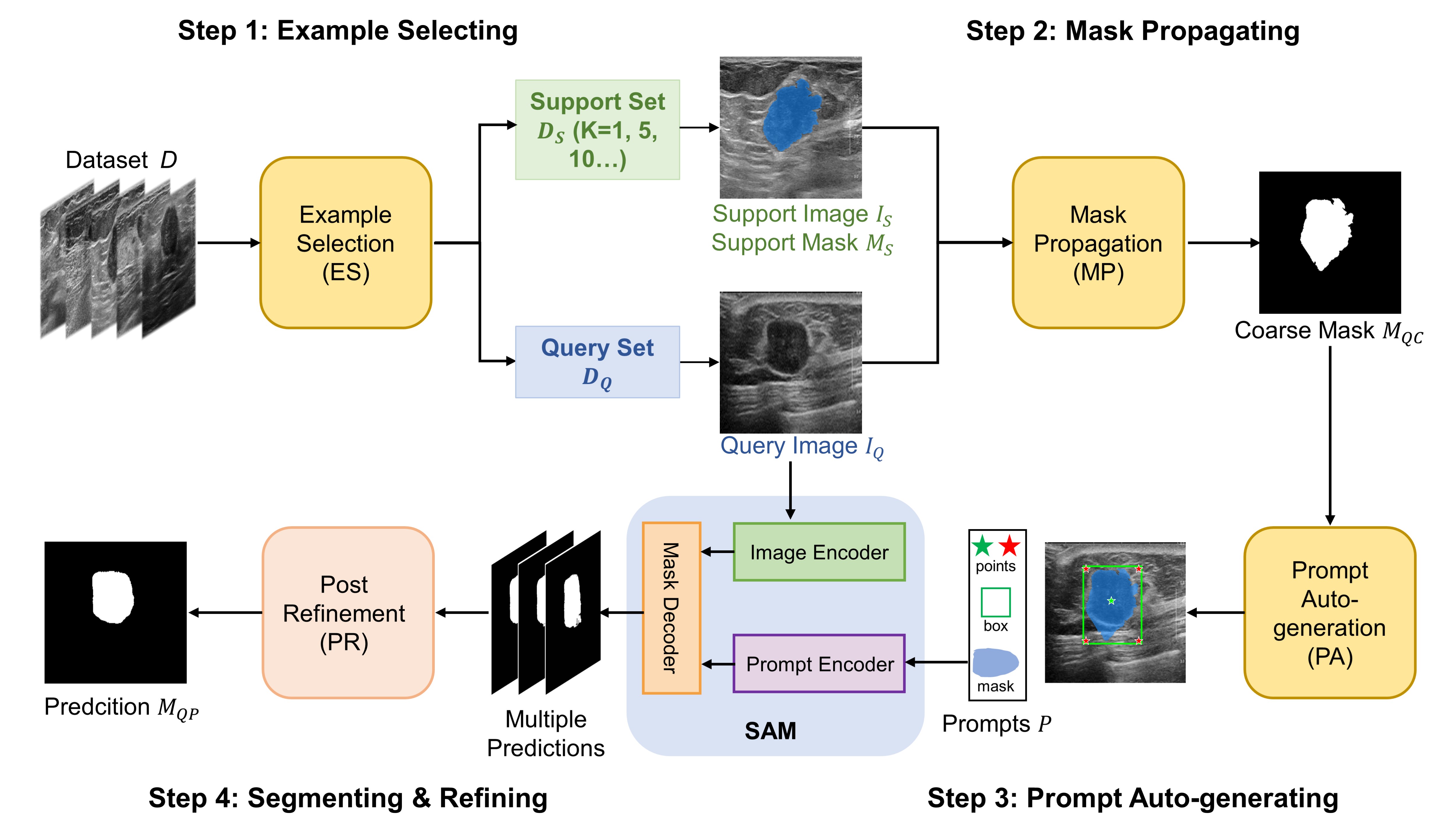}
\caption{Overview of proposed framework SAM-MPA.}
\label{fig2}
\end{figure}

As shown in Fig.\ref{fig2}, our proposed SAM-MPA framework comprises four primary steps: (1) Sample Selection: Seletcs $K$ samples($K=1, 5, 10$) from $D$ to form the support set $D_S$, with the remaining $N-K$ samples consituting the query set $D_Q$. (2) Mask Propagation: Transefers mask information from $D_S$ to $D_Q$ to generate coarse masks for all unlabeled images in $D_Q$. (3) Prompt Auto-generating: Automatically generates diverse prompts from the coarse masks. (4) Segmentation and Refining: Utilizes SAM to segment $D_Q$ based on the coarse masks and generated prompts, followed by a Post-Refinement process to further optimize the segmentation results. In the following sections, we will provide a detailed explanation of each step.


\subsection{Support Example Selection}
Samples with high similarity typically share more common feature points or structures, enabling better feature matching and precise transformation estimation. Thus our goal is to maximize the similarity between $D_S$ and $D_Q$. As computing similarity directly from raw images is expensive, we adopt a two-step approach: first extracting image features, then calculating similarity. Leveraging SAM's training on large-scale image datasets, its image encoder $Encoder_I$ possesses powerful representation capabilities. We utilize SAM for this process without the need to train a new model from scratch:
\begin{equation}
  \label{eq1}
  z_i = Encoder_I(d_i)   
\end{equation}
Using Eq.\ref{eq1}, we project the original dataset $D$ into feature space $Z=\{z_i\}_{i=1}^N$. For one-shot tasks ($K=1$), we select the sample most similar to others. Similarity is computed as:
\begin{equation}
    \label{eq2}
    Similarity(d_i) = \left( \frac{1}{N-1} \sum_{j=1, j\neq i}^N D(z_i,z_j) \right)^{-1} \\
\end{equation}
where $D(z_i,z_j)$ is the cosine distance between features: $D(z_i,z_j)=\frac{||z_i||\cdot ||z_j||}{||z_i|| \times ||z_j||}$. The final samle is determined as:
\begin{equation}
  \label{eq3}
  s_1 = \mathrm{argmax} \; Similarity(d_i)
\end{equation}

In few-shot tasks where $K > 1$, simply selecting the $K$ most similar samples may result in a support set $D_S$ that is overly concentrated in the feature space, failing to effectively cover the entire sample distribution. To address this, we employ the K-Centroid algorithm to partition the samples into $K$ clusters. The center of each cluster is designated as a support sample, collectively forming $D_S$, while the remaining samples within each cluster serve as query samples for their respective support sample. To enhance data coverage and mitigate the randomness inherent in K-Centroid's sensitivity to initial points, we utilize the K-Center-Greedy algorithm to determine the starting points. We initialize the K-Center-Greedy's process with $s_1$, the sample of highest similarity, rather than a random selection. This approach ensures consistent and representative results by eliminating variability due to random initialization.


\subsection{Mask Propagation}
After successfully constructing $D_S$ and $D_Q$, we employ a registration-based mask propagation module to generate coarse masks for samples in $D_Q$ based on the labeled data in $D_S$. Given the limited number of samples, particularly those with masks, we utilize an unsupervised B-spline elastic registration(BER) algorithm\cite{gu2014bidirectional}. Specifically, for a support image $s_i$ and its corresponding query image $q_j$, we compute a deformation field $\Phi$ using BER. We then apply $\Phi$ to the mask $m_i$ through a spatial transformer(ST) network, resulting in a coarse segmentation $mc_j$ for $q_j$, as shown in Eq.\ref{eq5}:
\begin{equation}
    \label{eq4}
    \Phi = BER(s_i, q_j)
\end{equation}
\begin{equation}
    \label{eq5}
    mc_j = ST(m_i, \Phi) = m_i \circ \Phi
\end{equation}

ST is a non-learnable neural network module designed for spatial transformations. It utilizes $\Phi$ to deform $m_i$ into $mc_j$. Specifically, ST employs a grid generation and sampling process to remap and interpolate each pixel in $m_i$ according to $\Phi$, thereby producing $mc_j$. This process can be described by Eq. \ref{eq6} and \ref{eq7}.
\begin{equation}
    \label{eq6}
    p^\prime = p + u(p)
\end{equation}
\begin{equation}
    \label{eq7}
    mc_j(p) = m_i \circ \Phi(p) = \sum_{q\in Z(p^\prime)} m_i(q) \prod_{d\in \{x,y\}} (1 - \left| p^\prime_d - q_d \right|)
\end{equation}
where $u(p)$ is the displacement of pixel $p$ in the deformation field $\Phi$. $Z(p^\prime)$ is the neighboring pixel of pixel $p^\prime$. $d$ traverses the space in all directions, where $x$ and $y$ represent the horizontal and vertical dimensions of the image, respectively.
\begin{figure}[H]
\centering
\includegraphics[width=0.5\textwidth]{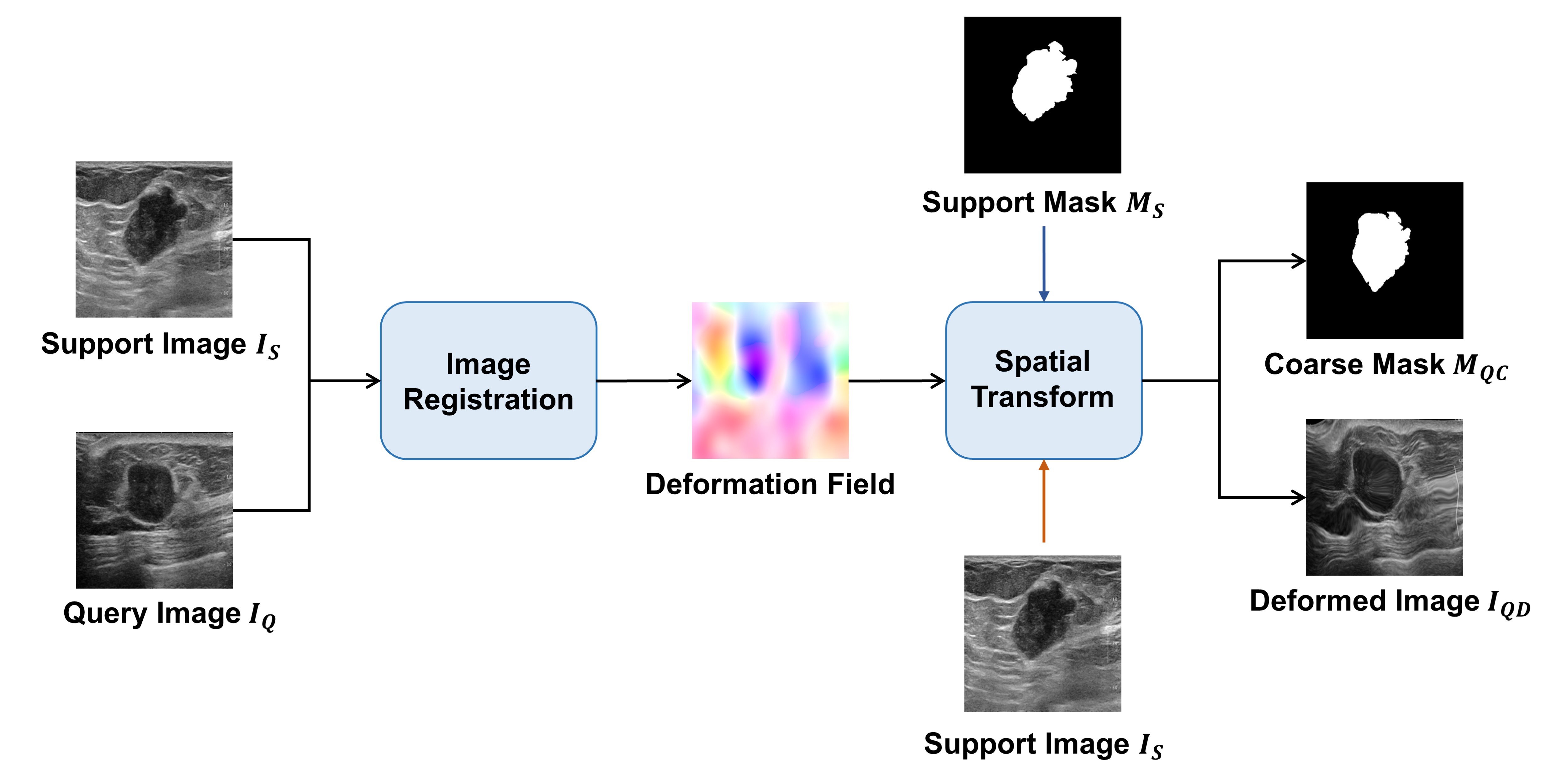}
\caption{The workflow of the Mask Propagation module.}
\label{fig3}
\end{figure}

\subsection{Prompts Auto-generation and Post Refinement Strategy}
Based on the coarse mask of the query image, we automatically generate various prompts required by SAM. As mentioned in Section \ref{sec2.2}, these prompts include point, box, and mask formats.The generation methods are as follows:
\begin{enumerate}[label=(\arabic*), leftmargin=*, align=left]  
  \item \textbf{Point prompts}: Foreground points are located at the centroid of the coarse segmentation, and background points are the four corners of the minimum bounding box surrounding the coarse segmentation.   
  \item \textbf{Box prompts}: The minimum bounding box of the coarse segmentation.  
  \item \textbf{Mask prompts}: Apply sigmoid softening to the coarse segmentation.
\end{enumerate}
We then feed the original image along with the generated prompts into SAM to produce an initial segmentation. To further optimize the output, we employ a Post-Refinement strategy that substitutes the preliminary segmentation for the old mask prompt and reprocess this updated input through SAM to yield a more precise delineation of the target object.
\begin{figure}[H]
\centering
\includegraphics[width=0.55\textwidth]{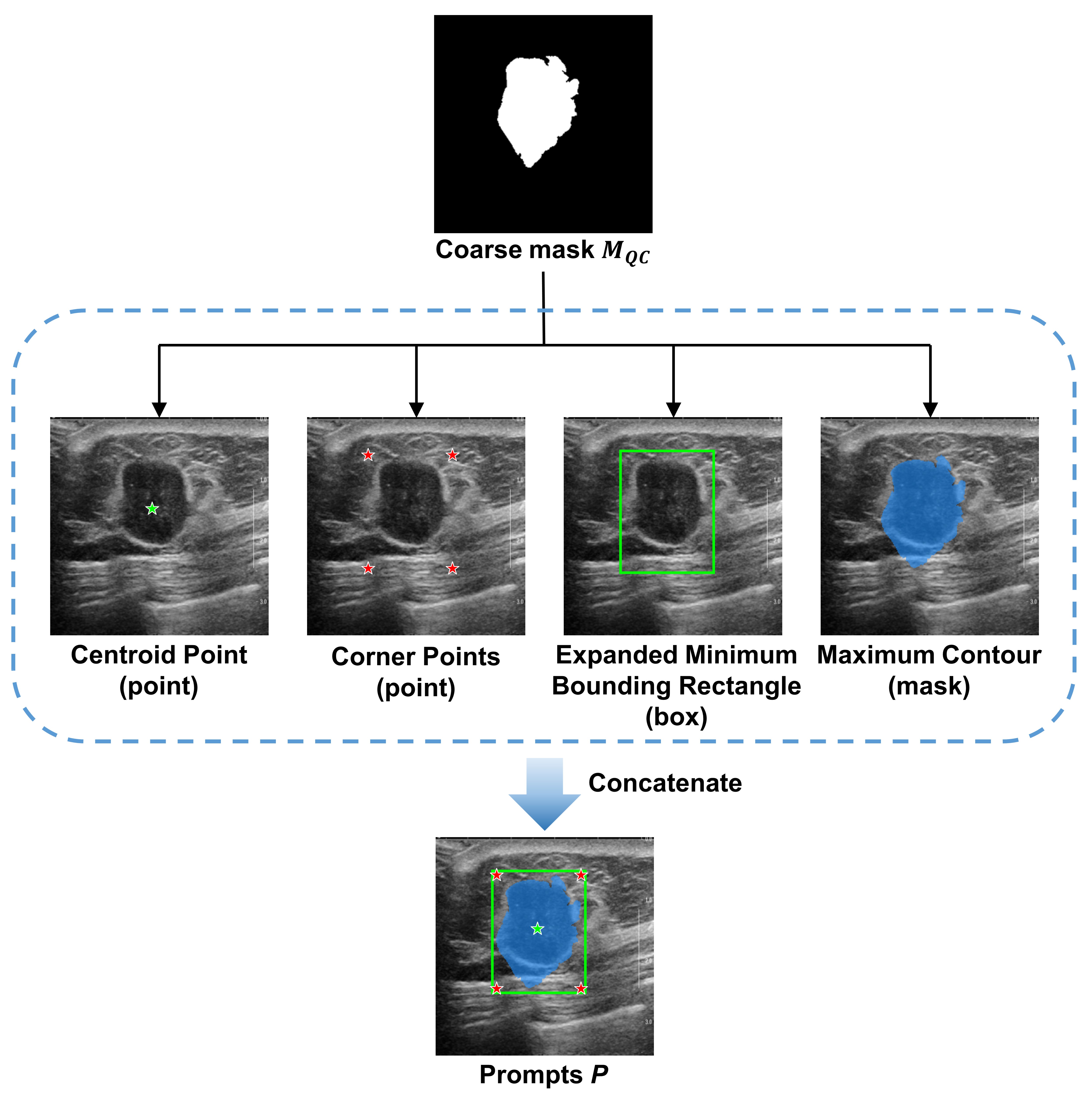}
\caption{The workflow of the Prompt Auto-generation module.}
\label{fig4}
\end{figure}


\section{Experiments}
\subsection{Datasets}
We evaluate our method on two medical image datasets with different modalities. Each image in the datasets has accurate pixel-wise segmentation labels that has been precisely annotated by an experienced radiologist as the ground truth. All images were resized to 256 × 256 pixels before segmentation.
\begin{itemize}[label=\textbullet, itemsep=0ex, leftmargin=*]  
\item 
    \textbf{Breast US:}\,Breast US is a private breast ultrasound dataset collected from Fudan University Shanghai Cancer Center, Shanghai, China. All images are obtained using the machine of the Aixplorer ultrasound system (SuperSonic Imagine S.A., Aix-en-Provence, France) at 7–15 MHz. It contains 200 images.
\item 
    \textbf{Chest X-Ray:}\, Chest X-ray is a standard digital image database for Tuberculosis is created by the National Library of Medicine, Maryland, USA in collaboration with Shenzhen No.3 People’s Hospital, Guangdong Medical College, Shenzhen, China\cite{jaeger2013automatic,candemir2013lung}. It contains 704 images with labels. The Chest X-rays are from out-patient clinics, and were captured as part of the daily routine using Philips DR Digital Diagnose systems.
\end{itemize}
\subsection{Implementation Details and Comparison Methods}
We evaluated the few-shot scenarios for 1, 5 and 10-shot settings. For sample selection and specific segmentation tasks, we employed the vit\_b model as SAM's image encoder. During the registration process, we utilized the B-Spline registration function from elastic-5.1.0, setting NumberOfResolutions to 3 while keeping other parameters at their default values.

The Dice coefficient served as our evaluation metric. All methods were implemented on the PyTorch platform, with computations accelerated using two Nvidia V100 32GB GPUs.

We compare the performances the manual prompting methods and automatic prompting methods based on SAM. The manual-prompting methods include using SAM with 1 point, 1 bounding box and the hybrid prompts (1 point and 1 bounding box). The auto-prompting few-shot method is PerSAM.



\section{Results}
\subsection{Ablation Study}
\label{Section5.1}
In this section, we conduct an ablation study to evaluate the proposed SAM-MPA method in a 10-shot setting, with results presented in Table \ref{Abla}. This systematic experiment assessed the contribution of each key component to the overall performance, specifically examining the Example Selection(ES) module, Mask Propagation(MP) module, Prompt Auto-generation(PA) module, and Post Refinement(PR) strategy. The results show that enabling each of the components leads to different levels of performance improvement, which intuitively reveals the functional importance of each component in the complete process.

In our ablation study, we carefully considered the interdependencies among components, establishing a sequential activation order. Mask Propagation (MP), the cornerstone of the SAM-MPA, was activated first. This component enables effective mask information transfer between support and query sets, without which the entire framework would be inoperable. Activating MP alone achieved a high-precision coarse segmentation with a Dice of 70.67\%, validating its efficacy in transmitting prior information. Next, we added Prompt Auto-generation (PA), which leverages the coarse masks produced by MP. The PA module, which automatically generates prompts for SAM based on the coarse masks, yielded a 0.83\% increase in Dice, validating its effectiveness. While MP and PA combined already enable SAM-based automatic segmentation, there was clearly room for enhancement. We then introduced the Example Selection (ES) component to assess how support set sample selection impacts segmentation results. The Dice increased by 2.34\%, demonstrating that our curated support samples provide superior guidance for query image segmentation. Finally, we activated the Post Refinement module, which replaces initial coarse masks with SAM-generated ones to enhance prompt quality. This step achieves a performance gain of 0.51\%, underscoring its role in boosting segmentation prediction performance.




\begin{table}[H]
\centering
\caption{Ablation study of the SAM-MPA on Breast US in the setting of 10-shot. ES: Example Selection; MP: Mask Propagation;	PA: Prompt Auto-generation;	PR: Post Refinement.}
\label{Abla}
\begin{tabular}{@{}cccccc@{}}
\toprule
ES & MP & PA & PR & Dice (\%) & Gain (\%) \\ \midrule
   & $\checkmark$  &    &    & 70.67     &           \\
   & $\checkmark$  & $\checkmark$  &    & 71.50     & \textbf{+0.83}     \\
$\checkmark$  & $\checkmark$  & $\checkmark$  &    & 73.84    & \textbf{+2.34}     \\
$\checkmark$  & $\checkmark$  & $\checkmark$  & $\checkmark$  & \textbf{74.53}     & \textbf{+0.69}     \\ \bottomrule
\end{tabular}
\end{table}

\subsection{Quantitative and Qualitative Results}

Table \ref{Table2} presents a performance comparison between SAM-MPA and other methods. The "Manual" approach refers to providing manual prompts to the original SAM model for direct target image segmentation, without involving few-shot settings such as support and query images. Results show that manually providing both point and bbox prompts yields the best performance, achieving dice scores of 70.32\% and 86.73\% on the two datasets. However, this method requires tedious manual confirmation of prompt points and bounding box drawing for each test image, such as processing 704 images in the ChestXray dataset. In contrast, our method achieves excellent segmentation performance by providing precise masks for 1, 5, or 10 support images. Using 10 masked support images, SAM-MPA achieves a dice score of 74.53\% on ultrasound images, surpassing the manual prompting method by 4.21\% and PerSAM by 18.93\%. For X-ray images, using just one support image yields a performance of 93.48\%, outperforming manual prompting method by 6.75\% and PerSAM by 46.55\%. These results demonstrate that our method not only avoids the cumbersome manual prompting process but also achieves satisfactory segmentation accuracy.
\begin{table}[H]
\caption{Quantitative comparison of Dice (\%) on Breast US and Chest X-ray in different K-shot settings. pt: point; bbox: bounding box.}
\label{Table2}
\resizebox{\textwidth}{!}{
\begin{tabular}{@{}cccccccc@{}}
\toprule
\multicolumn{2}{c}{\multirow{2}{*}{Method}}         & \multicolumn{3}{c}{Breast US} & \multicolumn{3}{c}{Chest X-ray} \\ \cmidrule(l){3-8} 
\multicolumn{2}{c}{}                                & 1-shot   & 5-shot  & 10-shot  & 1-shot   & 5-shot   & 10-shot   \\ \midrule
\multirow{3}{*}{Manual}    & SAM (1 point)          & \multicolumn{3}{c}{31.10}     & \multicolumn{3}{c}{50.42}      \\
                           & SAM (1 bbox)           & \multicolumn{3}{c}{67.43}     & \multicolumn{3}{c}{80.79}      \\
                           & SAM (1 point + 1 bbox) & \multicolumn{3}{c}{70.32}     & \multicolumn{3}{c}{86.73}       \\ \midrule
\multirow{2}{*}{Automatic} & PerSAM                 & 21.70    & 49.12   & 55.60    & 46.93    & 60.14    & 76.60    \\
                           & SAM-MPA(\textbf{Ours})                & \textbf{66.92}    & \textbf{68.65}   & \textbf{74.53}    & \textbf{93.48}    & \textbf{94.21}    & \textbf{94.36}     \\ \bottomrule
\end{tabular}}
\end{table}
\begin{figure}[H]
\centering
\includegraphics[width=0.85\textwidth]{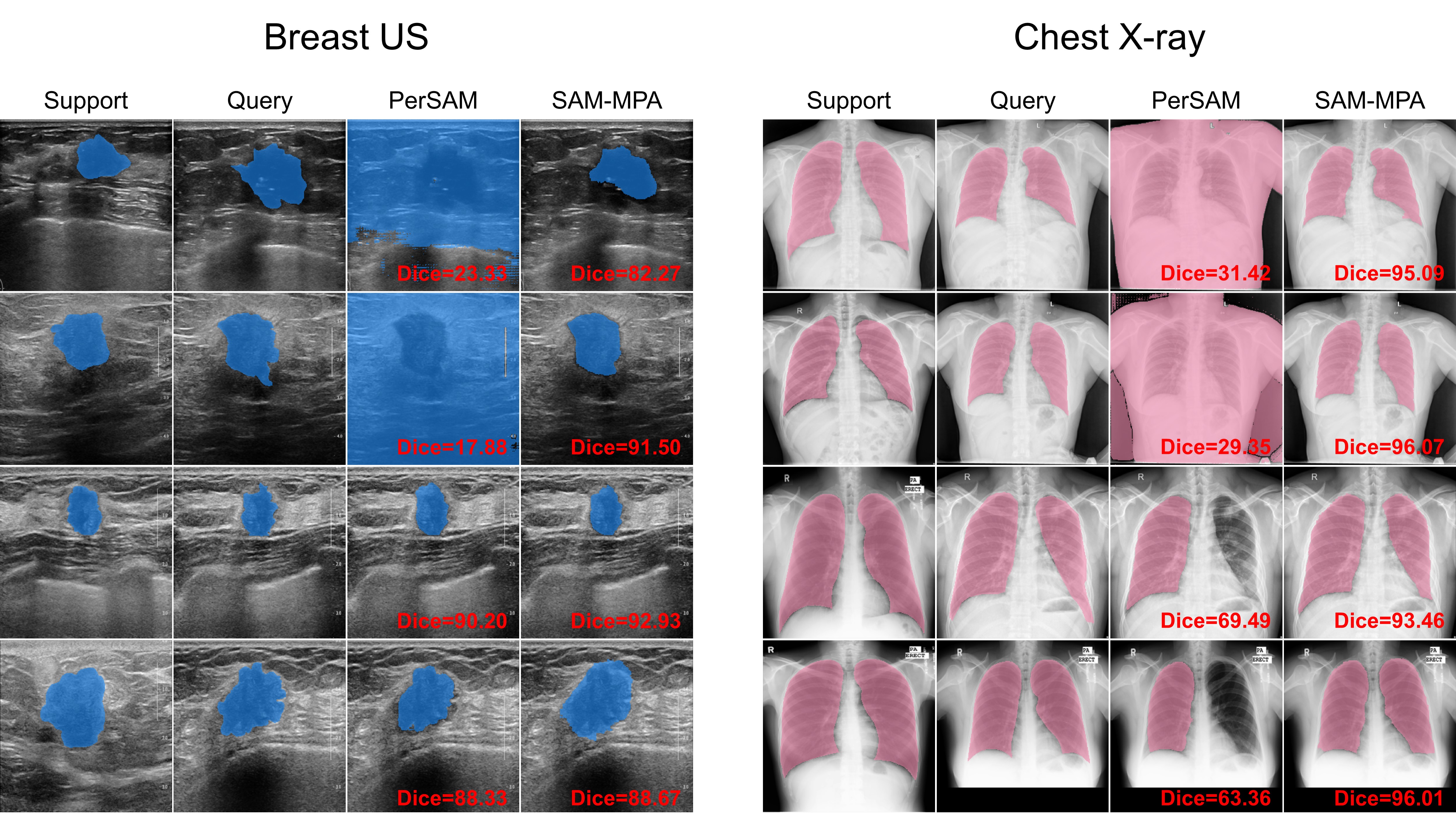}
\caption{Visual comparison of segmentation on Breast US and Chest X-ray in 10-shot setting.}
\label{fig5}
\end{figure}

The experimental results illustrated in Fig.\ref{fig5} demonstrate the segmentation performance of SAM-MPA and PerSAM across various query images. SAM-MPA consistently delivers satisfactory and stable segmentation results, exhibiting minimal performance variance across different images. In contrast, PerSAM's performance is characterized by significant fluctuations. While PerSAM achieves commendable segmentation results on certain images, it struggles considerably with others, sometimes failing to identify the correct target regions entirely. This inconsistency primarily stems from the subtle distinctions between foreground and background in medical images. The prototype similarity strategy employed by PerSAM proves ineffective in such scenarios, leading to instances where large portions of the image are erroneously classified as target regions. SAM-MPA, on the other hand, showcases superior performance. It adeptly identifies and accurately segments target regions, yielding satisfactory results even on images that pose challenges for PerSAM. This stark contrast underscores SAM-MPA's robustness and adaptability in handling complex medical images, highlighting its effectiveness in overcoming the limitations faced by alternative approaches.


\section{Discussion}



Few-shot methods in medical image segmentation often struggle with the need for large amounts of annotated data for pre-training. To address this issue, we propose the SAM-MPA framework, which leverages SAM for few-shot segmentation of medical images, effectively circumventing the demand for extensive labeled data. By automatically generating high-quality prompts for SAM, we successfully unlock its potential, achieving impressive segmentation results with just a handful of annotated samples (1, 5, or 10 masked images).

First, we will analyze the efficacy of each component in SAM-MPA. Ablation studies in Section \ref{Section5.1} demonstrate that every element plays an indispensable role. The sample selection module constructs a high-quality support set by selecting central samples from each cluster in the sample space and yields the most significant improvement(2.34\% performance gain). This underscores the importance of representative support samples in transferring label prior knowledge to query instances and highlights the sensitivity of foundation models like SAM to annotated example selection in downstream tasks. Then, the mask propagation module forms the backbone of the framework, facilitating efficient flow of annotation information between support and query samples. As evident from Fig.\ref{fig2}, SAM-MPA cannot function properly without this component. Employing unsupervised registration techniques, we achieve label propagation between support and query samples, attaining a substantial baseline segmentation performance of 70.67\%. Next, the automatic prompt generation component is equally crucial. Recent foundation models like SAM and GPT have shown that efficient prompts are key to unlocking a model's potential. Experimental results in Table\ref{Table2} reveal the significant impact of different prompts on SAM's performance. Using a single point prompt severely underperforms compared to box prompts, with performance gaps of 36.33\% and 30.27\% on US and Xray datasets, respectively. Our component automatically generates high-quality point, box, and mask prompts in one go, effectively enhancing segmentation performance. Lastly, the Post Refinement strategy leverages SAM's output mask as input for another round, optimizing segmentation results through SAM's inherent capabilities, proving to be an effective approach.

Second, we explore why our method outperforms others. While manually providing point and box prompts for images can yield decent results, this process becomes tedious when applied to a large quantity of images, requiring an enormous amount of manual labor. In contrast, SAM-MPA only requires manual masking of 1-10 images, with the rest of the process being fully automated. This significantly reduces human effort while achieving high-performance segmentation results, presenting a clear advantage. Compared to the current automated few-shot method PerSAM, our approach demonstrates substantial performance improvements. As shown in Table\ref{Table2}, with just one annotated image, our method outperforms PerSAM by 45.22\% on ultrasound images and 46.55\% on X-ray images. This is primarily due to: (1) The low contrast between foreground and background in medical images, which hinders PerSAM's region similarity-based strategy. (2) PerSAM's limited prompts (one foreground point and one background point) fail to fully harness SAM's potential. In contrast, SAM-MPA provides richer prompts, including one foreground point, four background points, a bounding box, and a coarse segmentation mask, offering SAM more comprehensive prior knowledge and effectively tapping into its capabilities.

Third, we briefly discuss the limitations of our method and future prospects. While SAM exhibits excellent generalization, it was pre-trained on natural images, resulting in a domain gap with medical images. Future work could explore foundation models specifically pre-trained on medical imaging data. Additionally, this study focused on 2D image applications, leaving 3D scenarios unexplored. Future research should investigate the feasibility and generalizability of SAM in segmenting 3D images.

\section{Conclusion}
Our SAM-MPA framework addresses the key challenges of adapting SAM to few-shot medical image segmentation including selecting representative examples, accurately propagating prior knowledge, and generating suitable prompts through dedicated modules. By leveraging mask propagation-based auto-prompting, we enhance SAM’s generalization capabilities for specific medical datasets. SAM-MPA demonstrates superior performance across diverse datasets in few-shot scenarios, offering a practical solution for high- accuracy segmentation with minimal user annotation. This work has the potential to significantly advance medical image analysis and diagnosis, providing a valuable tool for real-world healthcare  applications.
\\

\noindent \textbf{\large{Acknowledgments}} \\
This work was supported in part by the National Natural Science Foundation of China under Grant 62371139 and Grant 82227813; and in part by the Science and Technology Commission of Shanghai Municipality under Grant 22ZR1404800.

\end{document}